\documentclass[conference]{IEEEtran}
\IEEEoverridecommandlockouts

\usepackage{verbatim}
\usepackage{mathtools, arcs}
\usepackage[dvipsnames]{xcolor}
\usepackage{soul}
\usepackage{cite}
\usepackage{amsmath,amssymb,amsfonts}
\usepackage{algorithmic}
\usepackage{graphicx}
\usepackage{textcomp}
\usepackage{xcolor}
\usepackage{fancyhdr}
\def\BibTeX{{\rm B\kern-.05em{\sc i\kern-.025em b}\kern-.08em
    T\kern-.1667em\lower.7ex\hbox{E}\kern-.125emX}}

\usepackage{romannum}
\usepackage{amsmath}
\usepackage{tikz}
\usepackage{mathdots}
\usepackage{yhmath}
\usepackage{cancel}
\usepackage{color}
\usepackage{siunitx}
\usepackage{array}
\usepackage{multirow}
\usepackage{amssymb}
\usepackage{gensymb}
\usepackage{tabularx}
\usepackage{booktabs}
\usetikzlibrary{fadings}
\usetikzlibrary{patterns}
\usetikzlibrary{shadows.blur}
\usetikzlibrary{shapes}

\newcommand{\conftitle}{\footnotesize 2021 26$\textsuperscript{th}$ International Computer Conference, Computer Society of Iran (CSICC), Tehran, Iran}

\fancypagestyle{pageStyleOne}{%
    \fancyhf{}
    
    \fancyhead[C]{\conftitle}
    \fancyfoot[L]{\footnotesize 978-1-6654-1241-4/21/\$31.00 \copyright2021 {IEEE}}
}


\begin{document}
\title{Zero-Shot Estimation of Base Models' Weights in Ensemble of Machine Reading Comprehension Systems for Robust Generalization}

\author{\IEEEauthorblockN{Razieh Baradaran}
\IEEEauthorblockA{\textit{Technology and Engineering Department} \\
\textit{University of Qom}\\
Qom, Iran\\
r.baradaran@stu.qom.ac.ir}
\and
\IEEEauthorblockN{Hossein Amirkhani}
\IEEEauthorblockA{\textit{Technology and Engineering Department} \\
\textit{University of Qom}\\
Qom, Iran\\
amirkhani@qom.ac.ir}
}
\maketitle
\thispagestyle{pageStyleOne}

\pagestyle{empty}

\begin{abstract}
One of the main challenges of the machine reading comprehension (MRC) models is their fragile out-of-domain generalization, which makes these models not properly applicable to real-world general-purpose question answering problems. In this paper, we leverage a zero-shot weighted ensemble method for improving the robustness of out-of-domain generalization in MRC models. In the proposed method, a weight estimation module is used to estimate out-of-domain weights, and an ensemble module aggregate several base models' predictions based on their weights. The experiments indicate that the proposed method not only improves the final accuracy, but also is robust against domain changes.
\end{abstract}


\section{Introduction}
Machine Reading Comprehension (MRC) is one of the main components of today's open-domain question answering systems \cite{Drqa,yang-2019-end-open, angelidis-etal-2019-book}. Its aim is to answer the questions from the related context(s). One of the main issues in the MRC models is that they are highly dependent on their train dataset, and are fragile to domain changes so that their accuracy drops sharply in out-of-domain\footnote{In this paper, we do not distinguish conceptually between out-of-domain and out-of-distribution, like the MRQA shared task {\cite{fisch-etal-2019-mrqa}}} datasets \cite{talmor-berant-2019-multiqa}. However, in real-world question answering systems, it is necessary to be able to answer the questions from a wide range of domains with acceptable accuracy.

In recent years, some studies have focused on domain adaptation and knowledge transfer in the MRC models~\cite{b1,chung-etal-2018-supervised,Nishida-et-al-2020-unsupervised-LM,wang-etal-2019-adversarial}. But the aim of most of them is to adapt or transfer existing models' knowledge to a specific target domain, while having a domain-independent MRC model remains an unresolved issue. 

On the other hand, the common approach used in the most previous studies is supervised or semi-supervised transfer learning, which needs some data from the target domain. Even in the unsupervised domain adaptation approach, some raw texts from the target domain are available. However, in general-purpose question answering systems, due to the diversity of natural languages, you can not determine a specific domain as the target one. Therefore, the zero-shot setting, which assumes no data from the target domain is available, remains as an unexplored area.

In this paper, we investigate the generalization stability of MRC models in out-of-domain datasets and propose a simple zero-shot method to improve the generalization robustness of MRC models against domain changes. The proposed method uses an accuracy-based weighted ensemble, which includes several base models trained on separate datasets, a weight estimation module, and ensemble module to aggregate the base models' outputs on the target domain based on their out-of-domain estimated weights. In this method, there is no need for any target data in the training phase. 

The rest of this paper is organized as follows. In Section \Romannum{2}, the related work is reviewed. The proposed method is introduced in Section \Romannum {3}. The experiments are presented and discussed in Section \Romannum {4}, and the final section is dedicated to the conclusion and future work.  

\section{Related work}
The MRC task is a popular natural language processing task with several studies in recent years \cite{Drqa, van2020enhancing,hu2019read,liu2019neural}. Among these, some studies focus on generalization capability and transfer learning using supervised or unsupervised learning approaches across question answering (QA) or MRC models.
\subsection{Supervised Transfer Learning}
Chung et al. \cite{chung-etal-2018-supervised}, investigated the effect of transferring knowledge from one question answering model to other ones with different train datasets. They showed that pretraining a QA model on a source dataset and fine-tuning it on a target one can improve the performance of the target model. MultiQA \cite{talmor-berant-2019-multiqa} investigated the transfer and generalization capability of MRC models across different datasets. They showed that pretraining models on multiple datasets can reduce the need for a large amount of data from the target domain. They also stated that the MRC models have low generalizability in the zero-shot setting. So that, although training an MRC model on multiple datasets can lead to a more generalized model, it does not perform as well as the best model on the target dataset. Also, the MRQA shared task \cite{fisch-etal-2019-mrqa} focused on the generalization capability of the MRC models for out-of-domain data using 18 different datasets. They split these datasets into three parts, train, evaluation, and test, and explored various ideas to tackle the generalization problem in this task. This has been followed by multiple studies like \cite{Takahashi-etal-2019-cler,Wu2020ImprovingQG,Guo-etal-2020-multireQA}. Even though these studies use labeled target data for training a model, our focus is on the zero-shot setting where no data is available from the target domain. 
\subsection{Unsupervised Transfer Learning}
Some studies leveraged the unsupervised approach to transfer knowledge from a labeled source domain to an unlabeled target domain. In \cite{wang-etal-2019-adversarial}, an adversarial domain adaptation model is introduced, where the knowledge is transferred from a highly labeled source domain to an unlabeled target domain. Cao et al. \cite{b1} introduced a self-training structure for domain adaptation in MRC. This model used BERT\cite{Devlin-2018-BERT} to predict labels for target domain samples and filtered low confidence ones, and then trained target MRC model with an adversarial network. In \cite{Nishida-et-al-2020-unsupervised-LM}, a multi-task learning approach has been used to train both source domain MRC and target domain language model with shared layers. They showed this approach improved the accuracy in the target domain with only unlabeled passages.

In the mentioned studies, the aim is to improve performance on the specific target domain, while in our work, the aim is to have a general model that has robust performance on a wide range of datasets.

\section{Proposed Method}
\subsection{Machine Reading Comprehension}
Machine Reading Comprehension is a supervised learning task that learns to respond to the input questions from the related input context(s).

\begin{equation}
\begin{split}
f:(Q,C)\longrightarrow A\\ 
Q=(q_{1},q_{2},...,q_{l_{q}})\\ 
C=(c_{1},c_{2},...,c_{l_{c}})\label{eq1},
\end{split}
\end{equation}
where Q, C, A, $l_{q}$, and $l_{c}$ are the input question, input context, output answer, question length, and context length, respectively.

The output of the MRC model can be classified as selective or generative \cite{baradaran2020survey}. In the selective mode, the answer is an exact span of the input context, while in generative mode, the answer is a free form text. In this paper, we focus on the selective MRC models. The outputs of a selective MRC model are two probability distributions over context tokens for the start and end position of the answer. 

\subsection{Accuracy-Based Weighted Ensemble}
\begin{figure}[h]
\centering
\tikzset{every picture/.style={line width=0.75pt}} 

\begin{tikzpicture}[x=0.75pt,y=0.75pt,yscale=-1,xscale=1]

\draw   (199.8,28.79) -- (199.8,54.83) .. controls (199.8,57.91) and (180.1,60.41) .. (155.8,60.41) .. controls (131.5,60.41) and (111.8,57.91) .. (111.8,54.83) -- (111.8,28.79) .. controls (111.8,25.71) and (131.5,23.21) .. (155.8,23.21) .. controls (180.1,23.21) and (199.8,25.71) .. (199.8,28.79) .. controls (199.8,31.87) and (180.1,34.37) .. (155.8,34.37) .. controls (131.5,34.37) and (111.8,31.87) .. (111.8,28.79) ;
\draw   (364.3,28.6) -- (364.3,55.69) .. controls (364.3,58.9) and (344.6,61.5) .. (320.3,61.5) .. controls (296,61.5) and (276.3,58.9) .. (276.3,55.69) -- (276.3,28.6) .. controls (276.3,25.4) and (296,22.8) .. (320.3,22.8) .. controls (344.6,22.8) and (364.3,25.4) .. (364.3,28.6) .. controls (364.3,31.81) and (344.6,34.41) .. (320.3,34.41) .. controls (296,34.41) and (276.3,31.81) .. (276.3,28.6) ;
\draw   (104.65,91.7) .. controls (104.65,82.18) and (127.1,74.47) .. (154.8,74.47) .. controls (182.5,74.47) and (204.95,82.18) .. (204.95,91.7) .. controls (204.95,101.22) and (182.5,108.93) .. (154.8,108.93) .. controls (127.1,108.93) and (104.65,101.22) .. (104.65,91.7) -- cycle ;
\draw   (269,92.61) .. controls (269,83.09) and (291.45,75.38) .. (319.15,75.38) .. controls (346.85,75.38) and (369.3,83.09) .. (369.3,92.61) .. controls (369.3,102.13) and (346.85,109.84) .. (319.15,109.84) .. controls (291.45,109.84) and (269,102.13) .. (269,92.61) -- cycle ;
\draw   (422.8,115.37) -- (422.8,139.07) .. controls (422.8,141.87) and (403.21,144.15) .. (379.05,144.15) .. controls (354.89,144.15) and (335.3,141.87) .. (335.3,139.07) -- (335.3,115.37) .. controls (335.3,112.56) and (354.89,110.29) .. (379.05,110.29) .. controls (403.21,110.29) and (422.8,112.56) .. (422.8,115.37) .. controls (422.8,118.17) and (403.21,120.45) .. (379.05,120.45) .. controls (354.89,120.45) and (335.3,118.17) .. (335.3,115.37) ;
\draw   (177,144.77) -- (299.8,144.77) -- (299.8,181.21) -- (177,181.21) -- cycle ;
\draw    (335.8,131.2) -- (300.9,157.31) ;
\draw [shift={(299.3,158.51)}, rotate = 323.2] [color={rgb, 255:red, 0; green, 0; blue, 0 }  ][line width=0.75]    (10.93,-3.29) .. controls (6.95,-1.4) and (3.31,-0.3) .. (0,0) .. controls (3.31,0.3) and (6.95,1.4) .. (10.93,3.29)   ;
\draw   (137.8,209.98) .. controls (137.8,201.6) and (153.47,194.8) .. (172.8,194.8) .. controls (192.13,194.8) and (207.8,201.6) .. (207.8,209.98) .. controls (207.8,218.37) and (192.13,225.16) .. (172.8,225.16) .. controls (153.47,225.16) and (137.8,218.37) .. (137.8,209.98) -- cycle ;
\draw   (257.8,211.6) .. controls (257.8,203.43) and (273.47,196.8) .. (292.8,196.8) .. controls (312.13,196.8) and (327.8,203.43) .. (327.8,211.6) .. controls (327.8,219.78) and (312.13,226.4) .. (292.8,226.4) .. controls (273.47,226.4) and (257.8,219.78) .. (257.8,211.6) -- cycle ;
\draw   (90.8,59.48) .. controls (90.8,34.8) and (110.8,14.8) .. (135.48,14.8) -- (389.12,14.8) .. controls (413.8,14.8) and (433.8,34.8) .. (433.8,59.48) -- (433.8,193.52) .. controls (433.8,218.2) and (413.8,238.2) .. (389.12,238.2) -- (135.48,238.2) .. controls (110.8,238.2) and (90.8,218.2) .. (90.8,193.52) -- cycle ;
\draw   (93.62,282.06) .. controls (93.62,258.94) and (112.36,240.2) .. (135.48,240.2) -- (391.94,240.2) .. controls (415.06,240.2) and (433.8,258.94) .. (433.8,282.06) -- (433.8,407.62) .. controls (433.8,430.74) and (415.06,449.48) .. (391.94,449.48) -- (135.48,449.48) .. controls (112.36,449.48) and (93.62,430.74) .. (93.62,407.62) -- cycle ;
\draw   (316.8,253.76) -- (316.8,281.51) .. controls (316.8,284.73) and (291.73,287.34) .. (260.8,287.34) .. controls (229.87,287.34) and (204.8,284.73) .. (204.8,281.51) -- (204.8,253.76) .. controls (204.8,250.54) and (229.87,247.93) .. (260.8,247.93) .. controls (291.73,247.93) and (316.8,250.54) .. (316.8,253.76) .. controls (316.8,256.97) and (291.73,259.58) .. (260.8,259.58) .. controls (229.87,259.58) and (204.8,256.97) .. (204.8,253.76) ;
\draw   (110.8,322.28) .. controls (110.8,312.26) and (136.77,304.13) .. (168.8,304.13) .. controls (200.83,304.13) and (226.8,312.26) .. (226.8,322.28) .. controls (226.8,332.3) and (200.83,340.42) .. (168.8,340.42) .. controls (136.77,340.42) and (110.8,332.3) .. (110.8,322.28) -- cycle ;
\draw    (110.8,322.28) -- (226.8,322.28) ;
\draw   (286.8,320.14) .. controls (286.8,310.12) and (312.77,302) .. (344.8,302) .. controls (376.83,302) and (402.8,310.12) .. (402.8,320.14) .. controls (402.8,330.16) and (376.83,338.29) .. (344.8,338.29) .. controls (312.77,338.29) and (286.8,330.16) .. (286.8,320.14) -- cycle ;
\draw    (286.8,320.14) -- (402.8,320.14) ;
\draw   (199,346.82) -- (310.8,346.82) -- (310.8,381.97) -- (199,381.97) -- cycle ;
\draw   (196.2,412.28) .. controls (196.2,404.42) and (219.3,398.05) .. (247.8,398.05) .. controls (276.3,398.05) and (299.4,404.42) .. (299.4,412.28) .. controls (299.4,420.14) and (276.3,426.51) .. (247.8,426.51) .. controls (219.3,426.51) and (196.2,420.14) .. (196.2,412.28) -- cycle ;
\draw    (247.8,382.4) -- (247.8,396.05) ;
\draw [shift={(247.8,398.05)}, rotate = 270] [color={rgb, 255:red, 0; green, 0; blue, 0 }  ][line width=0.75]    (10.93,-3.29) .. controls (6.95,-1.4) and (3.31,-0.3) .. (0,0) .. controls (3.31,0.3) and (6.95,1.4) .. (10.93,3.29)   ;
\draw    (148.8,60.8) -- (148.8,74.8) ;
\draw [shift={(148.8,76.8)}, rotate = 270] [color={rgb, 255:red, 0; green, 0; blue, 0 }  ][line width=0.75]    (10.93,-3.29) .. controls (6.95,-1.4) and (3.31,-0.3) .. (0,0) .. controls (3.31,0.3) and (6.95,1.4) .. (10.93,3.29)   ;
\draw    (311.8,111.2) -- (281.18,143.35) ;
\draw [shift={(279.8,144.8)}, rotate = 313.6] [color={rgb, 255:red, 0; green, 0; blue, 0 }  ][line width=0.75]    (10.93,-3.29) .. controls (6.95,-1.4) and (3.31,-0.3) .. (0,0) .. controls (3.31,0.3) and (6.95,1.4) .. (10.93,3.29)   ;
\draw    (161.8,109.2) -- (205.23,143.56) ;
\draw [shift={(206.8,144.8)}, rotate = 218.35] [color={rgb, 255:red, 0; green, 0; blue, 0 }  ][line width=0.75]    (10.93,-3.29) .. controls (6.95,-1.4) and (3.31,-0.3) .. (0,0) .. controls (3.31,0.3) and (6.95,1.4) .. (10.93,3.29)   ;
\draw    (278.8,180.8) -- (299.15,194.67) ;
\draw [shift={(300.8,195.8)}, rotate = 214.29] [color={rgb, 255:red, 0; green, 0; blue, 0 }  ][line width=0.75]    (10.93,-3.29) .. controls (6.95,-1.4) and (3.31,-0.3) .. (0,0) .. controls (3.31,0.3) and (6.95,1.4) .. (10.93,3.29)   ;
\draw    (199.8,182.8) -- (174.63,193.99) ;
\draw [shift={(172.8,194.8)}, rotate = 336.03999999999996] [color={rgb, 255:red, 0; green, 0; blue, 0 }  ][line width=0.75]    (10.93,-3.29) .. controls (6.95,-1.4) and (3.31,-0.3) .. (0,0) .. controls (3.31,0.3) and (6.95,1.4) .. (10.93,3.29)   ;
\draw    (320.3,61.5) -- (319.32,73.38) ;
\draw [shift={(319.15,75.38)}, rotate = 274.74] [color={rgb, 255:red, 0; green, 0; blue, 0 }  ][line width=0.75]    (10.93,-3.29) .. controls (6.95,-1.4) and (3.31,-0.3) .. (0,0) .. controls (3.31,0.3) and (6.95,1.4) .. (10.93,3.29)   ;
\draw    (296.8,285.8) -- (323.09,301.76) ;
\draw [shift={(324.8,302.8)}, rotate = 211.26] [color={rgb, 255:red, 0; green, 0; blue, 0 }  ][line width=0.75]    (10.93,-3.29) .. controls (6.95,-1.4) and (3.31,-0.3) .. (0,0) .. controls (3.31,0.3) and (6.95,1.4) .. (10.93,3.29)   ;
\draw    (228.8,285.8) -- (192.6,303.33) ;
\draw [shift={(190.8,304.2)}, rotate = 334.15999999999997] [color={rgb, 255:red, 0; green, 0; blue, 0 }  ][line width=0.75]    (10.93,-3.29) .. controls (6.95,-1.4) and (3.31,-0.3) .. (0,0) .. controls (3.31,0.3) and (6.95,1.4) .. (10.93,3.29)   ;
\draw    (193,338.32) -- (214.92,346.13) ;
\draw [shift={(216.8,346.8)}, rotate = 199.6] [color={rgb, 255:red, 0; green, 0; blue, 0 }  ][line width=0.75]    (10.93,-3.29) .. controls (6.95,-1.4) and (3.31,-0.3) .. (0,0) .. controls (3.31,0.3) and (6.95,1.4) .. (10.93,3.29)   ;
\draw    (311.8,334.8) -- (291.52,346.78) ;
\draw [shift={(289.8,347.8)}, rotate = 329.41999999999996] [color={rgb, 255:red, 0; green, 0; blue, 0 }  ][line width=0.75]    (10.93,-3.29) .. controls (6.95,-1.4) and (3.31,-0.3) .. (0,0) .. controls (3.31,0.3) and (6.95,1.4) .. (10.93,3.29)   ;
\draw   (423.8,186.82) -- (423.8,210.52) .. controls (423.8,213.32) and (404.21,215.6) .. (380.05,215.6) .. controls (355.89,215.6) and (336.3,213.32) .. (336.3,210.52) -- (336.3,186.82) .. controls (336.3,184.02) and (355.89,181.74) .. (380.05,181.74) .. controls (404.21,181.74) and (423.8,184.02) .. (423.8,186.82) .. controls (423.8,189.63) and (404.21,191.9) .. (380.05,191.9) .. controls (355.89,191.9) and (336.3,189.63) .. (336.3,186.82) ;
\draw    (335.8,196.2) -- (301.46,173.31) ;
\draw [shift={(299.8,172.2)}, rotate = 393.69] [color={rgb, 255:red, 0; green, 0; blue, 0 }  ][line width=0.75]    (10.93,-3.29) .. controls (6.95,-1.4) and (3.31,-0.3) .. (0,0) .. controls (3.31,0.3) and (6.95,1.4) .. (10.93,3.29)   ;

\draw (120,34.89) node [anchor=north west][inner sep=0.75pt]  [font=\small] [align=left] {\begin{minipage}[lt]{48.12700000000001pt}\setlength\topsep{0pt}
\begin{center}
{\small $dataset_{1}$}
\end{center}

\end{minipage}};
\draw (281,37.69) node [anchor=north west][inner sep=0.75pt]   [align=left] {\begin{minipage}[lt]{48.12700000000001pt}\setlength\topsep{0pt}
\begin{center}
{\small $dataset_{n}$}
\end{center}

\end{minipage}};
\draw (224,32.62) node [anchor=north west][inner sep=0.75pt]  [font=\small] [align=left] {\textbf{. . .}};
\draw (108,82.72) node [anchor=north west][inner sep=0.75pt]  [font=\small] [align=left] {\begin{minipage}[lt]{63.92850000000001pt}\setlength\topsep{0pt}
\begin{center}
{\small $model_{1}$}
\end{center}

\end{minipage}};
\draw (270,81.96) node [anchor=north west][inner sep=0.75pt]  [font=\small] [align=left] {\begin{minipage}[lt]{63.92850000000001pt}\setlength\topsep{0pt}
\begin{center}
{\small $model_{n}$}
\end{center}

\end{minipage}};
\draw (338.3,121.37) node [anchor=north west][inner sep=0.75pt]  [font=\small] [align=left] {\begin{minipage}[lt]{58.5905pt}\setlength\topsep{0pt}
\begin{center}
{\small $dataset_{n+1}$}
\end{center}

\end{minipage}};
\draw (182,145.54) node [anchor=north west][inner sep=0.75pt]  [font=\small] [align=left] {\begin{minipage}[lt]{73.62700000000001pt}\setlength\topsep{0pt}
\begin{flushright}
weight estimation
\end{flushright}
\begin{center}
 module
\end{center}

\end{minipage}};
\draw (143,199.05) node [anchor=north west][inner sep=0.75pt]  [font=\small] [align=left] {\begin{minipage}[lt]{42.007000000000005pt}\setlength\topsep{0pt}
\begin{center}
{\small $weight_{1}$}
\end{center}

\end{minipage}};
\draw (264,202.05) node [anchor=north west][inner sep=0.75pt]  [font=\small] [align=left] {\begin{minipage}[lt]{42.007000000000005pt}\setlength\topsep{0pt}
\begin{center}
{\small $weight_{n}$}
\end{center}

\end{minipage}};
\draw (205,15.88) node [anchor=north west][inner sep=0.75pt]   [align=left] {{\small \textbf{train phase}}};
\draw (215,428.4) node [anchor=north west][inner sep=0.75pt]  [font=\small] [align=left] {\begin{minipage}[lt]{48.118500000000004pt}\setlength\topsep{0pt}
\begin{flushright}
\textbf{test phase}
\end{flushright}

\end{minipage}};
\draw (207,265.11) node [anchor=north west][inner sep=0.75pt]  [font=\small] [align=left] {\begin{minipage}[lt]{73.62700000000001pt}\setlength\topsep{0pt}
\begin{center}
{\small $dataset_{n'}$}
\end{center}

\end{minipage}};
\draw (118.85,320.53) node [anchor=north west][inner sep=0.75pt]  [font=\small] [align=left] {\begin{minipage}[lt]{63.92850000000001pt}\setlength\topsep{0pt}
\begin{center}
{\small $model_{1}$}
\end{center}

\end{minipage}};
\draw (297,318.61) node [anchor=north west][inner sep=0.75pt]  [font=\small] [align=left] {\begin{minipage}[lt]{63.92850000000001pt}\setlength\topsep{0pt}
\begin{center}
{\small $model_{n}$}
\end{center}

\end{minipage}};
\draw (137,305.09) node [anchor=north west][inner sep=0.75pt]  [font=\small] [align=left] {\begin{minipage}[lt]{42.007000000000005pt}\setlength\topsep{0pt}
\begin{center}
{\small $weight_{1}$}
\end{center}

\end{minipage}};
\draw (315,300.82) node [anchor=north west][inner sep=0.75pt]  [font=\small] [align=left] {\begin{minipage}[lt]{42.007000000000005pt}\setlength\topsep{0pt}
\begin{center}
{\small $weight_{n}$}
\end{center}

\end{minipage}};
\draw (198,353.18) node [anchor=north west][inner sep=0.75pt]  [font=\small] [align=left] {\begin{minipage}[lt]{75.16516000000001pt}\setlength\topsep{0pt}
\begin{flushright}
ensemble module
\end{flushright}

\end{minipage}};
\draw (202,401.99) node [anchor=north west][inner sep=0.75pt]  [font=\small] [align=left] {\begin{minipage}[lt]{61.9055pt}\setlength\topsep{0pt}
\begin{center}
final prediction
\end{center}

\end{minipage}};
\draw (225,81.2) node [anchor=north west][inner sep=0.75pt]  [font=\small] [align=left] {\textbf{. . .}};
\draw (218,201.81) node [anchor=north west][inner sep=0.75pt]  [font=\small] [align=left] {\textbf{. . .}};
\draw (241,312.2) node [anchor=north west][inner sep=0.75pt]  [font=\small] [align=left] {\textbf{. . .}};
\draw (339.3,191.82) node [anchor=north west][inner sep=0.75pt]  [font=\small] [align=left] {\begin{minipage}[lt]{58.5905pt}\setlength\topsep{0pt}
\begin{center}
{\small $dataset_{n+k}$}
\end{center}

\end{minipage}};
\draw (374.05,145.15) node [anchor=north west][inner sep=0.5pt]   [align=left] {\textbf{{\small .}}\\\textbf{{\small .}}\\\textbf{{\small .}}};
\end{tikzpicture}
\caption{The framework of the proposed method.}
\label{fig:framework}
\end{figure}
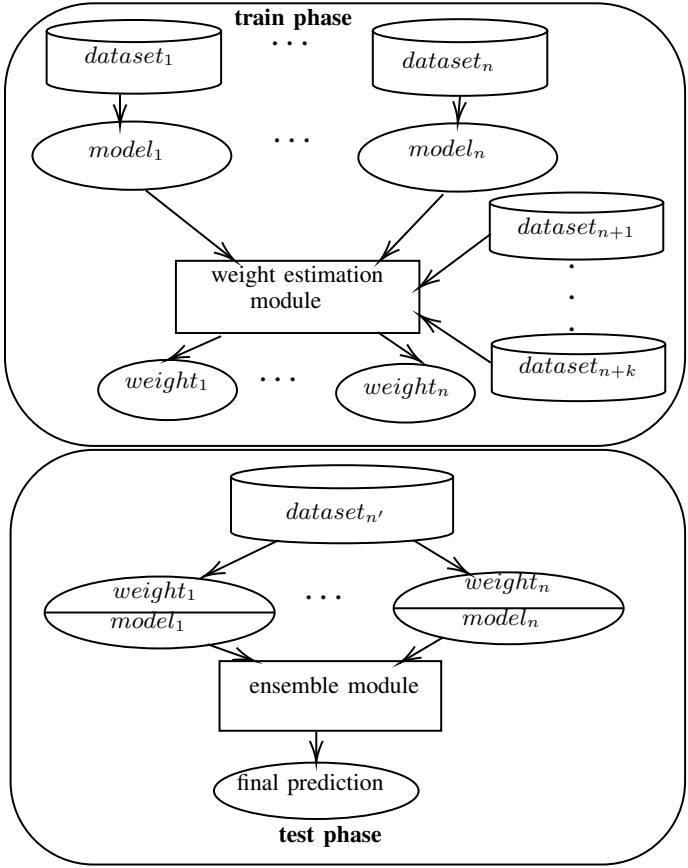

As stated in the previous section, most of the studies presented for domain adaptation in MRC task focus on transferring knowledge from the source domain to the desired target domain using some data from the target (at least raw passages). These models are fragile against domain changes and are not domain independent. In this study, we propose a simple zero-shot method to create a model which is robust against domain changes. This method, motivated by Large et al. work \cite{large2019probabilistic}, leverages several based models, a weight estimation module, and an ensemble module to generate the final prediction. The proposed framework is shown in Figure ~\ref{fig:framework}. Instead of adapting the model to the new domain, the proposed method uses the aggregation of several base model predictions, which is not only low cost compared to the previous approaches, but also can lead to more stability against domain changes.  

The base models have similar structures but are trained on separate training datasets ($dataset_{1}$ to $dataset_{n}$). In the weight estimation module, another set of datasets ($dataset_{n+1}$ to $dataset_{n+k}$) are used to estimate the accuracy of the base models as out-of-domain models' weights. In the test phase, the predictions for out-of-domain data ($dataset_{n'}$) are obtained using weighted ensemble of base models' predictions:
\begin{equation}
\hat{y}(dataset_{n'})=\sum_{j=1}^{n} w_{j}^{\alpha}\times \hat{y}_{{j}}(dataset_{n'})\label{eq2},
\end{equation}
where $w_{j}$ is the j-th model weight and $\hat{y}_{{j}}$ is the prediction of the $j$-th model on the target $dataset_{n'}$. The $\alpha$ is the only hyper-parameter of this method. The weight prediction module calculates $w_{j}$ as follows:
\begin{equation}
w_{j}=acc_{j}(\lbrace dataset_{n+1},...,dataset_{n+k}\rbrace)\label{eq3},
\end{equation}
where the $acc_{j}$ is the accuracy of the $j$-th model on the set of samples from $dataset_{n+1}$ to $dataset_{n+k}$.


\section{Experiments}
\subsection{Datasets}

\begin{table*}[htbp]
\caption{The datasets used in our experiments}
\begin{center}
\begin{tabular}{|c|c|c|c|c|c|}
\hline

&\textbf{Dataset} & \textbf{\textit{Qustion}}& \textbf{\textit{Context}} &\textbf{\textit{Train}}& \textbf{\textit{Test}} \\
\hline
\multirow{5}{*}{\Romannum{1}}&SQuAD&Crowdsourced&Wikipedia&86588&10507 \\
\cline{2-6}
&NewsQA&Crowdsourced &News& 74160 & 4212\\
\cline{2-6}
&Natural Questions&Search logs&Wikipedia&104071&12836 \\
\cline{2-6}
&DROP&Crowdsourced &Wikipedia&77409$^{\mathrm{a}}$&1503 \\
\cline{2-6}
&DuoRC&Crowdsourced& Movie plots& 60721$^{\mathrm{a}}$&1501\\
\hline
\multirow{3}{*}{\Romannum{2}}&TriviaQA&Trivia&Web snippets& 61688&7785\\
\cline{2-6}
&HotpotQA&Crowdsourced&Wikipedia&72928&5904 \\
\cline{2-6}
&SearchQA& Jeopardy&Web snippets&117,384&16980\\
\hline
\multirow{4}{*}{\Romannum{3}}&RACE&Domain experts&Examinations&-&674\\
\cline{2-6}
&TextbookQA& Domain experts&Textbook &-&1503\\
\cline{2-6}
&BioASQ&Domain experts&Science articles&-&1504\\
\cline{2-6}
&RelationExtraction&Synthetic&Wikipedia&-&2948\\
\hline
\multicolumn{5}{l}{$^{\mathrm{a}}$MRQA shared task does not have these train sets, so we used them }\\
\multicolumn{5}{l}{from the MultiQA project \cite{talmor-berant-2019-multiqa}} 
\end{tabular}
\label{tab-dataset}
\end{center}
\end{table*}

In this paper, we use 12 MRC datasets with different sizes and domains. The detailed information is shown in Table \ref{tab-dataset}. The datasets are configured corresponding to the MRQA shared task \cite{fisch-etal-2019-mrqa}. These datasets are split into three groups which are respectively used for the base model learning, the weight estimation, and testing the final model. 

\subsection{Experimental Results}
We performed three sets of experiments in this study. First, we trained base models with the simple and popular MRC model, BiDAF \cite{Seo2017BiDAF} on group \Romannum{1} datasets, and evaluated their in-domain and out-of-domain accuracies. The AllenNLP library\footnote{https://github.com/allenai/allennlp-reading-comprehension} is used for training the base models. As you can see in Table \ref{tab-group1}, in almost all datasets, the best results are obtained when the source and target domains are the same; and accuracy drops significantly for the unseen datasets. The used evaluation measure is F1 score which calculates the weighted average of the precision and recall between the predicted answer and ground-truth at the work level. 

\begin{table}[ht!bp]
\caption{The F1 score of the BiDAF model trained and tested on different datasets. The Natural Questions dataset is abbreviated as NQ.}
\begin{center}
\begin{tabular}{|c|c|c|c|c|c|}
\hline

\textbf{train} $\downarrow$ \textbf{test}$\rightarrow$ &\textbf{SQuAD} & \textbf{NewsQA}& \textbf{NQ} &\textbf{DROP}& \textbf{DuoRC} \\
\hline
\textbf{SQuAD}&77.83&43.89&34.86&5.79&39.45 \\
\hline
\textbf{NewsQA}&53.22 &51.12& 31.72& 11.44&32.88\\
\hline
\textbf{NQ}&38.83&23.88&66.04&13.88&19.90 \\
\hline
\textbf{DROP}&17.41&10.65&8.40&79.90&9.98 \\
\hline
\textbf{DuoRC}&37.06&22.98&7.10&5.91&35.48\\
\hline
\end{tabular}
\label{tab-group1}
\end{center}
\end{table}

The next experiment, presented in Table~\ref{tab-group2}, investigates the proposed accuracy-based weighted ensemble method which ensembles the base models' outputs according to Equation~\ref{eq2}. Each of the datasets from group \Romannum{2} (unseen during training) is used as a train set in the weight estimation module. For simplicity and speed considerations, we only used a subset of 5000 train samples from each dataset. 
The $\alpha$ hyper-parameter is chosen from 1 to 4, where the out-of-fold predictions over the train set are used as the validation set. We also compare this method with the fine-tuning approach, where the best base model is fine-tuned on the unseen datasets and evaluated on other datasets to measure its out-of-domain accuracy.
As shown in Table \ref{tab-group2}, the proposed weighted ensemble method obtains the highest accuracies. Specially, the results show that the fine-tuning approach cannot improve the model's performance for the unseen datasets. The simple ensemble method in this table is the arithmetic mean of the base models' outputs. 

\begin{table*}[htbp]
\caption{The F1 score of the base models, the fine-tuning approach, the simple ensemble method, and the proposed weighted ensemble method.}
\begin{center}
\begin{tabular}{|c|c|c|c|c|}
\hline
model&\textbf{train} $\downarrow$\textbf{ test}$\rightarrow$ &\textbf{HotpotQA} & \textbf{SearchQA}&\textbf{TriviaQA}\\
\hline
\multirow{5}{*}{\textbf{base models}}&\textbf{SQuAD}&43.30&19.27&40.94\\
&\textbf{NewsQA}&37.31&14.69&31.79\\
&\textbf{NQ}&20.10&13.35&18.04\\
&\textbf{DROP}&12.28&2.56&12.55\\
&\textbf{DuoRC}&29.63&14.43&28.72\\
\hline
\multirow{3}{*}{\textbf{fine-tune}}&\textbf{HotpotQA}&-&18.15&35.49\\
&\textbf{SearchQA}&36.42&-&36.99\\
&\textbf{TriviaQA}&36.43&21.63&-\\
\hline
\textbf{simple ensemble}&-&42.41&14.84&40.27\\
\hline
\multirow{3}{*}{\textbf{weighted ensemble}}&\textbf{HotpotQA}&-&\textbf{21.97}&\textbf{43.40}\\
&\textbf{SearchQA}&45.81&-&43.26\\
&\textbf{TriviaQA}&\textbf{46.23}&19.95&-\\
\hline
\end{tabular}
\label{tab-group2}
\end{center}
\end{table*}

\begin{table*}[ht!]
\caption{The F1 score of the base models, the fine-tuning approach, and the proposed weighted ensemble method trained on a combination of all datasets of group \Romannum{2} and tested on each dataset of group \Romannum{3}.}
\begin{center}
\begin{tabular}{|c|c|c|c|c|c|}
\hline
model&\textbf{train} $\downarrow$\textbf{ test}$\rightarrow$ &\textbf{RACE} & \textbf{TextbookQA}&\textbf{BioASQ}&\textbf{RelationExtraction}\\
\hline
\multirow{5}{*}{\textbf{base models}}&\textbf{SQuAD}&26.90&33.19&36.40&64.21\\
&\textbf{NewsQA}&21.75&23.03&24.58&44.96\\
&\textbf{NQ}&15.32&21.36&16.93&39.77\\
&\textbf{DROP}&6.47&3.24&10.51&13.97\\
&\textbf{DuoRC}&9.93&14.39&8.40&23.18\\
\hline
\textbf{fine-tune}&\textbf{Group \Romannum{2}}&17.71&22.63&26.80&60.14\\
\hline
\textbf{weighted ensemble}&\textbf{Group \Romannum{2}}&\textbf{27.15}&\textbf{34.69}&\textbf{38.68}&\textbf{65.39}\\
\hline
\end{tabular}
\label{tab-group3}
\end{center}
\end{table*} 

The last but not the least experiment explores the effect of simultaneous usage of multiple datasets (group \Romannum{2}) for weight estimation in the proposed method as well as in the fine-tuning approach. The performance of the methods are tested on the group \Romannum{3} datasets. To this end, we randomly select 5000 samples from each dataset in group \Romannum{2}, resulting in a total of 15000 samples, and estimate models' weights using the proposed module. The results are shown in Table \ref{tab-group3}. Despite using multiple datasets, the accuracy of the fine-tuning approach is still fragile for the unseen datasets, while the weighted ensemble method remains stable. 

According to the experiments, despite its simplicity, the proposed weighted ensemble method can obtain robust results in the out-of-domain data. It seems that the lack of a direct dependence on one particular data distribution is one of the factors contributing to its robustness. 
In addition, the proposed method does not need a high volume of data to estimate the weight of base models compared to training or fine-tuning an MRC model with many parameters. All these show the high capability of the proposed method in robust generalization on unseen data, and can be a start point for future research in this area.

\section{Conclusion and future work}
In this paper, we investigated the generalization capability of MRC models on out-of-domain datasets and proposed a simple weighted ensemble method to robustly generalize on the unseen datasets. We compared the robustness of our method with the fine-tuning approach, in which the base models are fine-tuned on one or multiple out-of-domain datasets and tested on other ones. 

In experiments, we used 5 base models trained on separate datasets. In the first step, we explored the generalization capability of the base models on out-of-domain datasets. The experimental results indicated the poor performance of MRC models on out-of-domain datasets. Then, we evaluated the proposed method and fine-tuning approach trained on one or multiple unseen datasets. The results indicated that our method leads to more robust generalization, where its accuracy on unseen datasets were better than all of the base models and fine-tuning approach.

The advantages of the proposed method are its simple implementation, its flexibility in adding new base models, and the lack of a direct dependence on the specific test data, which leads to more stable results on out-of-domain data.

For future work, we want to explore more sophisticated methods for the weighting module, such as sample-based weighting, which considers each input sample features for weighting different base models.



\end{document}